\def\year{2023}\relax
\documentclass[letterpaper]{article} 
\usepackage{aaai23}  
\usepackage{times}  
\usepackage{helvet}  
\usepackage{courier}  
\usepackage[hyphens]{url}  
\usepackage{graphicx} 
\usepackage{natbib}  
\usepackage{caption} 
\usepackage{algorithm}
\usepackage{algorithmic}
\usepackage{amsmath,amssymb}
\DeclareMathOperator{\E}{\mathbb{E}}

\usepackage{newfloat}
\usepackage{listings}
\floatstyle{ruled}
\newfloat{listing}{tb}{lst}{}
\floatname{listing}{Listing}

\nocopyright 

\title{A Generative Adversarial Network for Climate Tipping Point Discovery (TIP-GAN)}
\author {
    Jennifer Sleeman,\textsuperscript{\rm 1}
    David Chung, \textsuperscript{\rm 1}
    Anand Gnanadesikan, \textsuperscript{\rm 2}
    Jay Brett,\textsuperscript{\rm 1}
    Yannis Kevrekidis,\textsuperscript{\rm 2}
    Marisa Hughes,\textsuperscript{\rm 1}
    Thomas Haine ,\textsuperscript{\rm 2}
    Marie-Aude Pradal,\textsuperscript{\rm 2} 
    Renske Gelderloos ,\textsuperscript{\rm 2}
    Chace Ashcraft, \textsuperscript{\rm 1}
    Caroline Tang, \textsuperscript{\rm 3}
    Anshu Saksena,\textsuperscript{\rm 1}
    Larry White \textsuperscript{\rm 1}
}
\affiliations {
    \textsuperscript{\rm 1} Johns Hopkins University Applied Physics Laboratory\\
    \textsuperscript{\rm 2} Johns Hopkins University\\
    \textsuperscript{\rm 3} Duke University\\
    jennifer.sleeman@jhuapl.edu, david.chung@jhuapl.edu, chace.ashcraft@jhuapl.edu, jay.brett@jhuapl.edu,
    gnanades@jhu.edu, yannisk@jhu.edu,marisa.hughes@jhuapl.edu,
    thomas.haine@jhu.edu,pradalma@gmail.com,rgelder2@jhu.edu,
    caroline.tang@duke.edu,anshu.saksena@jhuapl.edu,larry.white@jhuapl.edu
}
\begin{document}

\maketitle
\begin{abstract}
We propose a new Tipping Point Generative Adversarial Network (TIP-GAN) for better characterizing potential climate tipping points in Earth system models. We describe an adversarial game to explore the parameter space of these models, detect upcoming tipping points, and discover the drivers of tipping points. In this setup, a set of generators learn to construct model configurations that will invoke a climate tipping point. The discriminator learns to identify which generators are generating each model configuration and whether a given configuration will lead to a tipping point. The discriminator is trained using an oracle (a surrogate climate model) to test if a generated model configuration leads to a tipping point or not. We demonstrate the application of this GAN to invoke the collapse of the Atlantic Meridional Overturning Circulation (AMOC). We share experimental results of modifying the loss functions and the number of generators to exploit the area of uncertainty in model state space near a climate tipping point. In addition, we show that our trained discriminator can predict AMOC collapse with a high degree of accuracy without the use of the oracle. This approach could generalize to other tipping points, and could augment climate modeling research by directing users interested in studying tipping points to parameter sets likely to induce said tipping points in their computationally intensive climate models.
\end{abstract}

\section{Introduction}
The concept of a climate tipping point was introduced in the climate research community decades ago.  Lenton et al. \cite{lenton2011early} describe a tipping point as a small change in forcing that leads to a large non-linear response which changes the state of the affected dynamical system, in this case a climate subsystem. Earth's geological record shows many examples where apparently small and steady changes in the tilt of the earth or concentrations of greenhouse gasses in the atmosphere result in sharp shifts in climate and ecosystems \citep{lenton2013environmental}. This has resulted in significant concern regarding whether current increases in atmospheric greenhouse gasses will produce similar tipping points in the near future. In 2018, the Intergovernmental Panel for Climate Change summarized in a special report the potential risks surrounding climate tipping points \cite{portner2019ocean}. Research related to this topic has highlighted a number of specific phenomena that could contribute to irreversible change to our world including retreat of ice, loss of forest cover, and changes in ocean currents \cite{lenton2019climate}.  

However, a problem with understanding these tipping points is that they are not robust across the Earth System models used to project future climates. This uncertainty arises in part because many processes in the model have to be simplified through the use of idealized representations and parameterizations. Differences in how individual processes are parameterized can result in tipping points occurring at different levels of global warming \cite{bahl2020scaling} complicating efforts to set ”safe" levels of greenhouse gasses.  Due to the large number of computations involved in calculating circulations and processes in multiple domains, running climate model simulations requires high performance computing environments and days if not weeks to complete.  If we want to use large climate models to explore potential tipping points, a brute-force search of the possible parameter space is simply not possible.  

We describe a methodology based on a Generative Adversarial Network (GAN) \cite{goodfellow2014generative}, and show how it could be used to direct climate researchers to areas in the search space that could reduce the number of climate simulations needed to be performed and enable faster climate tipping point discovery. Our approach, which was introduced in Sleeman et al. \cite{sleeman_aaai_fall_symposium_2022}, as part of a new methodology for performing climate tipping scientific discovery, the Tipping Point Generative Adversarial Network (TIP-GAN) acts as an AI assistant directing the climate modeler to areas of interest to explore scientifically, and generalizes in two ways:  1.) to other types of tipping points beyond climate, and 2.) to other types of scientific discovery problems.  Our described TIP-GAN approach offers a new way to explore abrupt changes in state space of dynamical systems.

The typical GAN architecture consists of a discriminator and generator deep neural network.  The two networks engage in an adversarial game based on the minimax algorithm.  The discriminator learns how to classify samples from a real distribution and learns how to distinguish real samples from fake samples that are generated by the generator.  The generator generates these samples randomly initially, but then learns from the discriminator which of its random samples do a better job at confusing the discriminator.  As these networks engage in this back and forth interaction, they both learn how to improve their part of the game with the goal of eventually reaching a Nash equilibrium.  GANs are well known for performing image generation  \cite{radford2015unsupervised} and text generation \cite{zhang2017adversarial}.  More recent exploration of novel use of GANs includes evolutionary GAN \cite{wang2019evolutionary}.  Our extension to this typical GAN setup that is novel is we introduce the concept of a surrogate model that acts as the oracle to the discriminator.  We also show how using multiple generators and a custom loss function which incorporates the concept of uncertainty can be used to find these abrupt changes in state space.

In this study, we applied TIP-GAN to a well-known tipping point, the Atlantic Meridional Overturning Circulation (AMOC). We use a simplified model of the AMOC which nonetheless involves a significant number of parameters and thus represents a high-dimensional search space to generate an extensive dataset of simulations. We then describe how TIP-GAN focuses on the parts of this search space that are close to a tipping point.
 
\section{Background - The AMOC as a Tipping Point}

The AMOC is a large-scale circulation pattern in which relatively warm, salty water flows into the North Atlantic, where it releases heat to the atmosphere, causing it to become denser and sink into the deep ocean. The AMOC helps make the Northern Hemisphere significantly warmer than the Southern Hemisphere and helps to ensure the habitability of Northern Europe. It also pushes the boundary between wet and moist climates north of the equator in Africa and South America \citep{zhang2005simulated}. Additionally it plays a key role in sequestering carbon dioxide in the deep ocean, and thereby indirectly regulates the global greenhouse effect and Earth's mean temperature \citep{marinov2008impact}. A collapse of the AMOC could have devastating effects on food security \cite{benton2020running}, rising sea levels \cite{bakker2022ocean}, Arctic related effects \cite{liu2022interaction} and vulnerable ecosystems \cite{velasco2021synergistic}. Because of this there is a strong sense of urgency in understanding, anticipating, and if possible, preventing a permanent AMOC shift.

A classic paper \cite{stommel1961thermohaline} suggested that the AMOC was very close to a tipping point today. Because the ability of the atmosphere to hold water vapor is strongly dependent on temperature (increasing by 7\% per degree C), the exchange of warm and cold air between the tropics and high latitudes transports freshwater polewards, making the high latitudes fresh-and therefore lighter-with respect to the tropics. Stommel developed a simple box model of the overturning and showed that it was subject to collapse when the density anomaly produced by freshening was 50\% of that produced by cooling and that this could be produced by a 10\% increase in freshwater flux. However, this model neglected a large number of feedbacks that stabilize the overturning. A series of expansions building on this model include \citet{gnanadesikan1999pycnocline,johnson2007reconciling,jones2016interbasin}. Feedbacks include: slowing the formation of dense water in the northern hemisphere causes less-dense water to ``accumulate" in the tropics, increasing the driving pressure gradient. Eddies in the ocean, which are analogous to weather systems in the atmosphere, stir fluid between polar and tropical regions, allowing the freshwater dumped in polar regions to escape. However, these eddies also allow the light water accumulating in the tropics to escape to the Southern Ocean, and thus act as a destabilizing feedback. There are significant gaps in our understanding of these processes. A more complex box model \citep{gnanadesikan2018flux} that includes these feedbacks suggests that instead of 10\%, the actual increase in freshwater flux required could be closer to 60\%.

This result is more consistent with the climate models, which generally simulate a relatively slow and steady decrease of the overturning in response to changes in greenhouse gasses \citep{weaver2012stability}. However, it is possible that these models have been systematically biased away from being too close to tipping points, as model configurations that tip in preindustrial control simulations or historical runs due to natural variability will generally be rejected as unrealistic during the development process.  It was thus quite notable when \citet{jackson2018hysteresis} found such a tipping point in a modern climate model with ultrahigh resolution in the ocean. This result raises the question of whether previous failures to find such tipping points were due to deficiencies the representation of ocean processes.

In this work we explore this question using the simplified box model of \cite{gnanadesikan2018flux} acting as a surrogate climate model in concert with the GAN.  The simplified model contains over twenty values of initial conditions and parameters. In some cases these values are only weakly constrained by observations and theory. Our long-term goal is to explore this high-dimensional space for cases that look a lot like the present day AMOC, but that are actually very close to a tipping point. Below, we present a proof-of-concept of how a GAN could be used to focus effort on cases near a tipping point.

\subsection{Tipping Point Formalized}

\begin{figure} 
\centering
\includegraphics[width=.8\columnwidth]{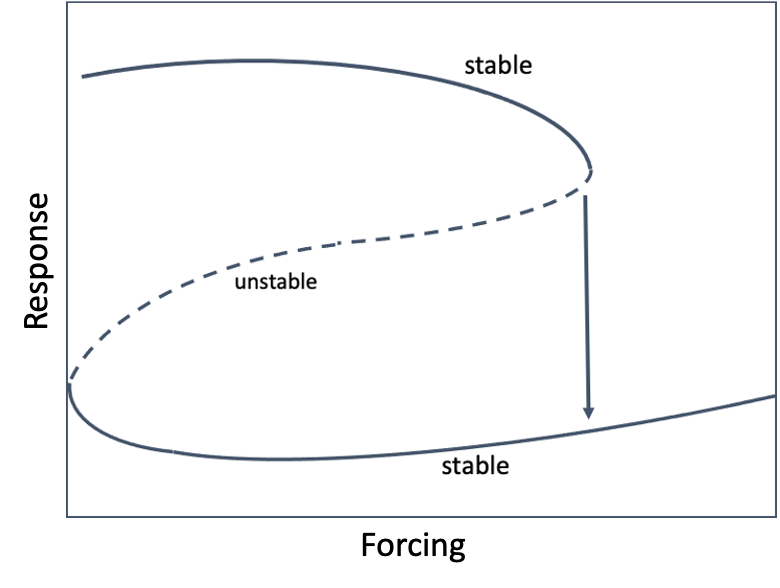}
\caption{Fold Bifurcation.}
\label{fig:fold}
\end{figure}

Identifying tipping points can be described more formally in terms of bifurcation and nonlinear dynamical systems \cite{thompson1990nonlinear}.  Typically established numerical bifurcation/continuation algorithms are used to discover the locus of ”hard" bifurcations that are known to underpin model tipping points. The most basic of these bifurcations is the saddle-node (fold) bifurcation, depicted in Figure \ref{fig:fold}, which is the easiest to identify. The \citet{stommel1961thermohaline} box model is known to be characterized by such a bifurcation. In a fold bifurcation, a range of values of a forcing (say the freshwater flux) can be associated with multiple values of an equilibrium response (say overturning). However, not all of these equilibrium states are stable- as illustrated in Figure \ref{fig:fold}, states lying along the dashed line can collapse to either the top or bottom solid lines if they are perturbed. The dashed line forms a {\em separatrix} between the basins of attraction of the two stable states. Thus a key goal of characterizing how close one is to the tipping point is to identify this separatrix.

\section{Related Work}
Though this is a relatively new area of research, there has been some early work related to using deep learning for early warning signal detection.  Work by Bury et al. \cite{bury2021deep} applied deep learning for early warning signal detection.  Also taking the approach of exploiting the dynamics and used a convolutional LSTM architecture to learn a prediction of the new states, focusing on behavior near the tipping point. The authors of that work propose that training the LSTM on the dynamics would enable the network to generalize to other types of models.  The task they wished to achieve with this method had a different objective than our method.  

In work by Deb et al. \cite{deb2022machine} another deep learning method was proposed for early warning detection that also used an LSTM and was focused on detecting state transitions. Again the objectives of this method were to perform a generalized early warning detection using the trained model on different classes of problems. 

Lapeyrolerie et al. \cite{lapeyrolerie2021teaching}  points out that the critical slowing down approach (which is used by the other related work) is problematic because the detection of these slowing down patterns (related to the bifurcations) are too general, leading to false negatives.  This further supports the premise that this area of research is still early in terms of applying deep learning to this problem.  

We propose that our method could be used an assistive AI to direct climate modelers to areas of the parameter space that warrant further investigation.  Rather than building a deep learning network that could generalize to different dynamical systems, it is the TIP-GAN machinery (the combination of the generators, discriminator, and the surrogate) that is generalizable to other systems, as other surrogates could be used, and the architecture is built to support different types of tipping point problems.  We also believe that the TIP-GAN learned latent space could be supported by an explainability component.  

\section{TIP-GAN}
\begin{figure} 
\centering
\includegraphics[width=1\columnwidth]{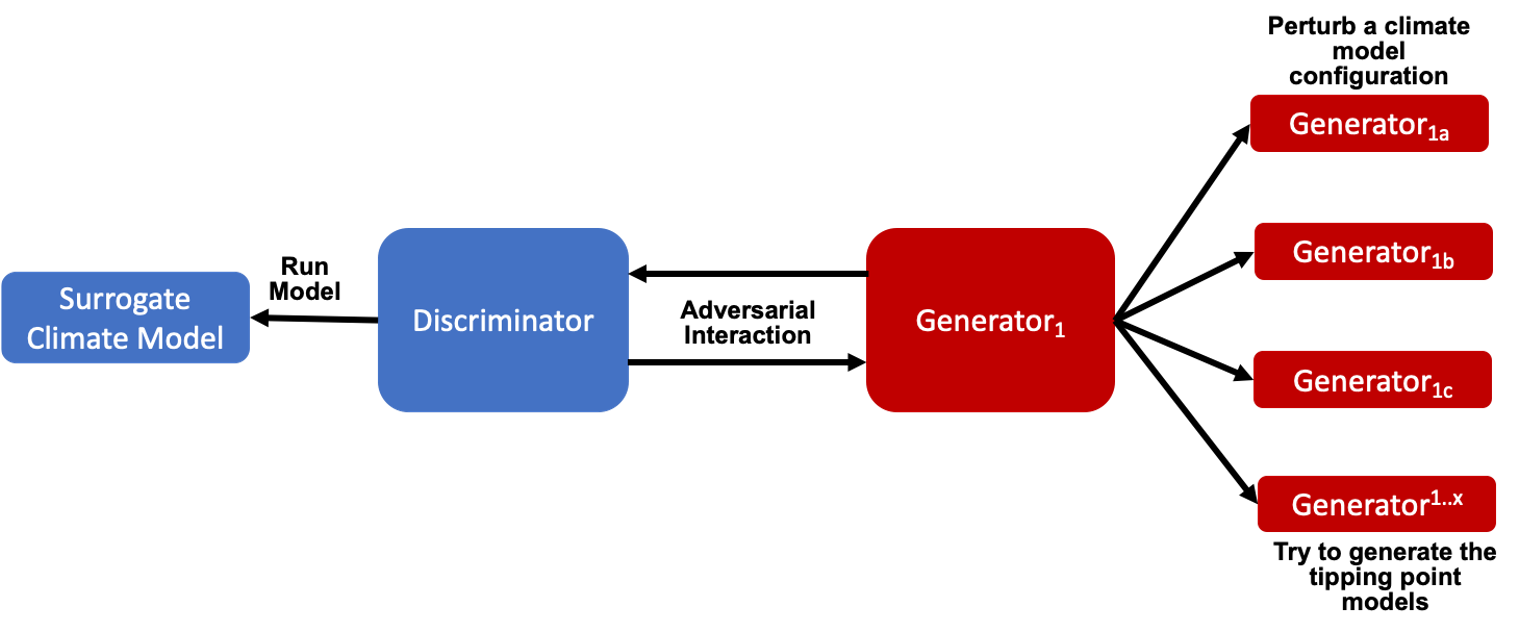}
\caption{The Climate Tipping Point GAN (TIP-GAN).}
\label{fig:gan}
\end{figure}

To overcome the challenges related to discovering tipping points in large climate simulations, we developed a novel, tipping point GAN architecture.  As depicted in Figure \ref{fig:gan}, TIP-GAN includes a discriminator, a set of generators, and a surrogate model.  The surrogate model acts as the oracle executing model configurations suggested by the generators.  

We use multiple generators to explore the different modalities of the distribution and to improve stability of the GAN.  Previous work using multiple generators \cite{hoang2018mgan,ghosh2018multi} showed that having multiple generators enabled the GAN to be more stable (reducing mode collapse common among GANs) during training as each generator tends to explore a different modality of the distribution (as diverse modes have been shown to combat mode collapse).  In our work we take advantage of this feature of using multiple generators with the idea that each generator would exploit a different modality of the tipping point parameter space.  However, our work shows new emergent behavior that results from having multiple generators.  We describe this behavior in our experimental results.

The TIP-GAN architectural approach acts as a general machinery for exploring tipping points, where different surrogate models could be used as the oracle and the problem setup is based on a parameterization for the model configuration.  The adversarial nature of GANs is a good model for this problem using the multi-generator setup to seek out the areas in state space where there are abrupt changes. After the discriminator is trained, it could be used as a classifier to predict whether a model configuration would result in an AMOC ”on" (AMOC non-collapse) or ”off" (AMOC collapse) state.

\section{TIP-GAN for AMOC Tipping Point Discovery}
We describe using TIP-GAN for AMOC tipping point discovery using a four box model as a means for developing the dataset, ground truth, and evaluation by comparing the results of TIP-GAN with the original results \cite{gnanadesikan2018flux} of experiments using the same three dimensional parameter space.  

\begin{figure}
\centering
\includegraphics[width=1\columnwidth]{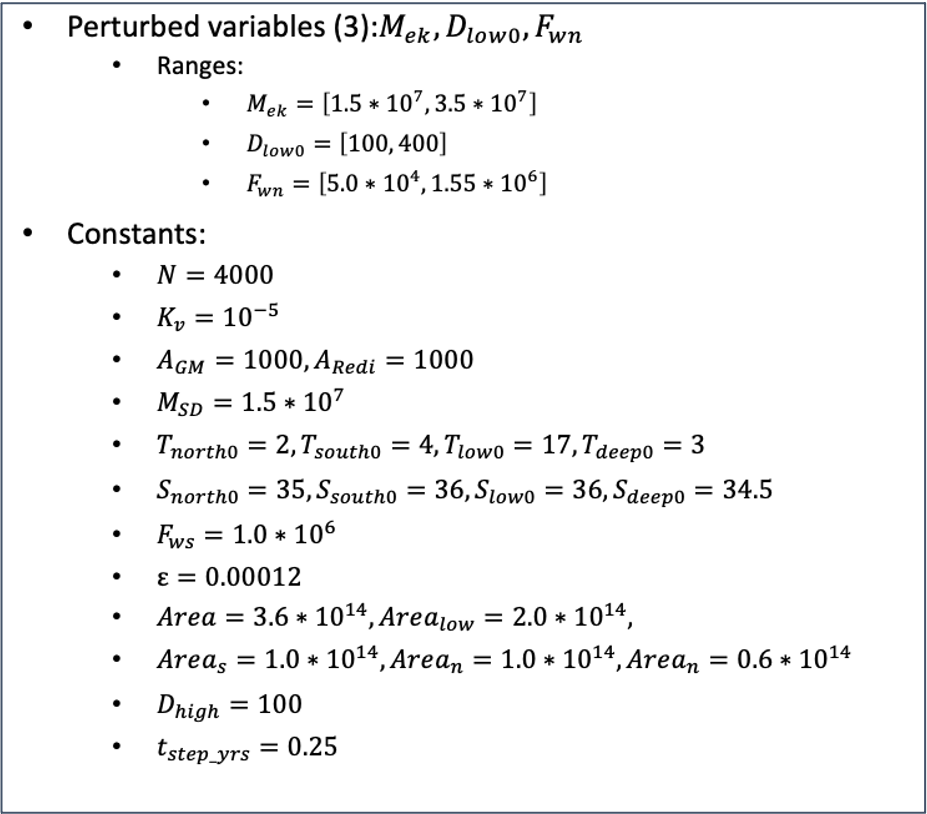}
\caption{Four box model experimental configuration replicated in TIP-GAN.}
\label{fig:setup}

\end{figure}

The TIP-GAN discriminator is trained to learn to predict which model configurations (See Figure \ref{fig:setup} for the model configuration) result in an AMOC state of ”on" or ”off".  In this four box model, this is equivalent to measuring the overturning variable $M_n$ and detecting when that variable changes from a positive value to a negative value. 

\begin{figure} 
\centering
\includegraphics[width=1\columnwidth]{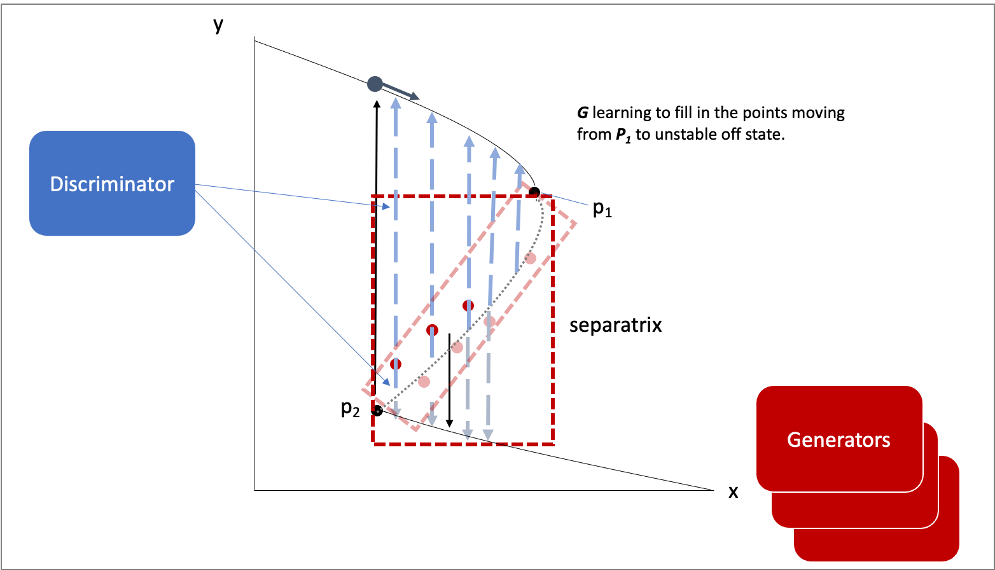}
\caption{A Two Dimensional Representation of the GAN Learning the area of the seperatrix.  G represents the generator, $P_1$ represents the point where the stable (on) state ends, and $P_2$ represents the point where the stable (off) state begins.}
\label{fig:gan_sep}
\end{figure}

The exploratory generators generate model configurations that include initial conditions and randomly selected model parameter values.  They are trained to identify the area where this change in state occurs, depicted in Figure \ref{fig:gan_sep} using two dimensions.  The solid lines indicate stable states and the dotted line indicate an unstable state, which is boxed by a red rectangle, the area which the generators are trained to learn. It is this area that the discriminator, when it predicts ”on" and ”off" states, is likely to be uncertain.  In state space this region is described in terms of the separatrix of a fold bifurcation.  Other types of bifurcations could be explored in terms of this generative model, however we focus on the fold bifurcation in this study.

The surrogate model is the four box model that we run to test a model configuration.  It acts as the oracle for the discriminator as it provides the actual ”on/off" labels for model configurations.

\subsection{Discriminator Objectives}
Given a configuration, the discriminator has two objectives:  
\begin{enumerate}
\item Identify the origin of the configuration (i.e. which generator produced it or if it was sampled from the real data distribution).
\item Correctly predict if the configuration will induce a shutoff state.
\end{enumerate}

At each update step, the discriminator will achieve these two objectives for $m(n+1)$ configurations where $m$ samples are obtained per each of $n$ generators and the additional $+1$ batch from the real data distribution.  Ground-truth shutoff labels are determined by consulting the surrogate model.

\subsection{Generator Objectives}
Using $n$ Generators, where $for \:i=1,...,n$, Generator $G_i$ will produce $m$ batch size configurations for the surrogate model to execute.  The generated configurations are passed through the discriminator to compute both the GAN logits and the AMOC state classification logits.  Each generator has two objectives: 

\begin{enumerate}

\item Guide the discriminator into predicting that its configurations are sampled from the real data distribution. 
\item Generate model configurations where the discriminator is least certain about the output state (i.e. AMOC shutoff vs. non-shutoff).
\end{enumerate}

\subsection{GAN Objective Formalized}
The objective can be defined in terms of an extension to the Multi-Agent Diverse Generative Adversarial Networks (MAD-GAN) \cite{ghosh2018multi} objective where:

\begin{equation}
\begin{aligned}
L_{MAD} &=\underset{\theta}{min} \underset{\phi}{max}\:V(G_\theta,D_\phi)\\
&=\E_{x~p_d}[log D_\phi(x)] + \E_{z~p_z}[log(1-D_\phi(G_\theta(z)))]
\end{aligned}
\end{equation}

In addition to maximizing the MAD-GAN objective, the discriminator is optimized to classify a configuration as either stable or unstable. The generators are optimized to produce configurations that the discriminator is most uncertain. The objective for this classification problem can be formalized as:

\begin{equation}
\begin{aligned}
L_{CLF} &=\underset{\theta}{min} \underset{\phi}{max}V(G_\theta,D_\phi)\\
&=-\E_{(x,y)~p_d}[y \: log \: D_\phi(x)+(1-y)(1-log \:D_\phi(x))]\\ 
&- 0.5 \: \E_{z~p_z}[log \: D_\phi(G_\theta(z))+(1-log \: D_\phi(G_\theta(z)))]
\end{aligned}
\end{equation}
\\
Combining the two objectives results in the TIP-GAN objective:
\begin{equation}
\begin{aligned}
L_{TIP}=L_{MAD} + L_{CLF}
\end{aligned}
\end{equation}

\subsection{The Surrogate Model}
\begin{figure} 
\centering
\includegraphics[width=.9\columnwidth]{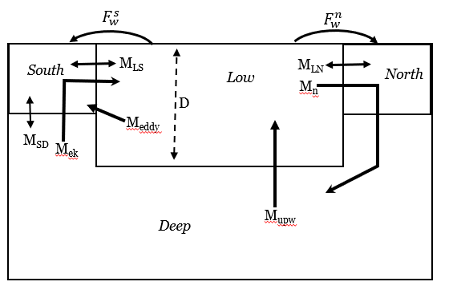}
\caption{The Four Box Model.}
\label{fig:fourbox}
\end{figure}

As previously noted, many of the dynamical processes involved in setting up the AMOC can be represented in simple box models \citep{levermann2010atlantic}, allowing for a much more extensive exploration of parameter space than in full Earth System Models. The box model we use here is taken from \cite{gnanadesikan2018flux} and is shown in Figure \ref{fig:fourbox}. It includes four boxes- a single box representing the deep ocean and three surface boxes representing the Southern Ocean, low-latitudes and North Atlantic/Arctic. The depth of the low-latitude box $D_{low}$ is determined by a mass balance equation, with the AMOC removing mass from the box when it is in its ”on" state and recycling it to the low latitudes when it is in its ”off" state. The mass transport associated with this is denoted as $M_n$ in Figure \ref{fig:fourbox}.  This removal is balanced by diffusively-driven upwelling in the low latitudes ($M_{upw}$)  and wind-driven upwelling in the Southern Ocean ($M_{ek}$). Eddy mixing in the Southern Ocean drives a flux of mass into the Southern Ocean $M_{eddy}$, representing an alternative pathway for converting non-dense water to dense water. Additionally, mixing fluxes ($M_{sl},M_{nl}$)  exchange tracers between the surface boxes and between the surface Southern Ocean box and the deep ocean. This last flux $M_s$ represents the formation of Southern Ocean deep water. All the fluxes except for $M_{ek}$ and $M_{s}$ depend on $D_{low}$. The magnitude of these fluxes is represented in units of Sverdrups (Sv), where Sv=1 million m$^3$/s, where 1 Sv is roughly equivalent to all the world's rivers combined.  While the mean values of $M_n$ have been estimated from multiple lines of observation to lie between 15 and 20 Sv \citep{broecker1998much,mcdonagh2015continuous}, the other fluxes are much less well constrained. For example, recent work found the upwelling flux $M_{ek}$ to vary between 13 and 33 Sv in modern climate models \citep{almeida2021impact}.  

In addition to the depth of the low latitude pycnocline $D_{low}$, temperatures and salinities are predicted in all four boxes (giving us nine equations with nine potential initial conditions). In the three surface boxes, temperatures are restored towards some equilibrium temperature and are thus only weakly responsive to changes in the overturning. Salinities are affected by atmospheric transports of freshwater $F_w^{n,s}$ which act to make low latitudes salty and high latitudes fresh.  These fluxes are much smaller than those associated with the overturning, with recent estimates for the North Atlantic lying between 0.17 and 0.57 Sv  \citep{mcdonagh2015continuous}. However, they can still produce large impacts on the salinity and density gradients- in the base case of \cite{gnanadesikan2018flux} an instantaneous increase in the flux from 0.55 Sv to 0.77 Sv was sufficient to collapse the AMOC. A common way of representing the tipping point of the AMOC is to plot the overturning transport $M_n$ as a function of the freshwater flux $F_w^n$.

\section{Experimental Setup}
\begin{figure} 
\centering
\includegraphics[width=1\columnwidth]{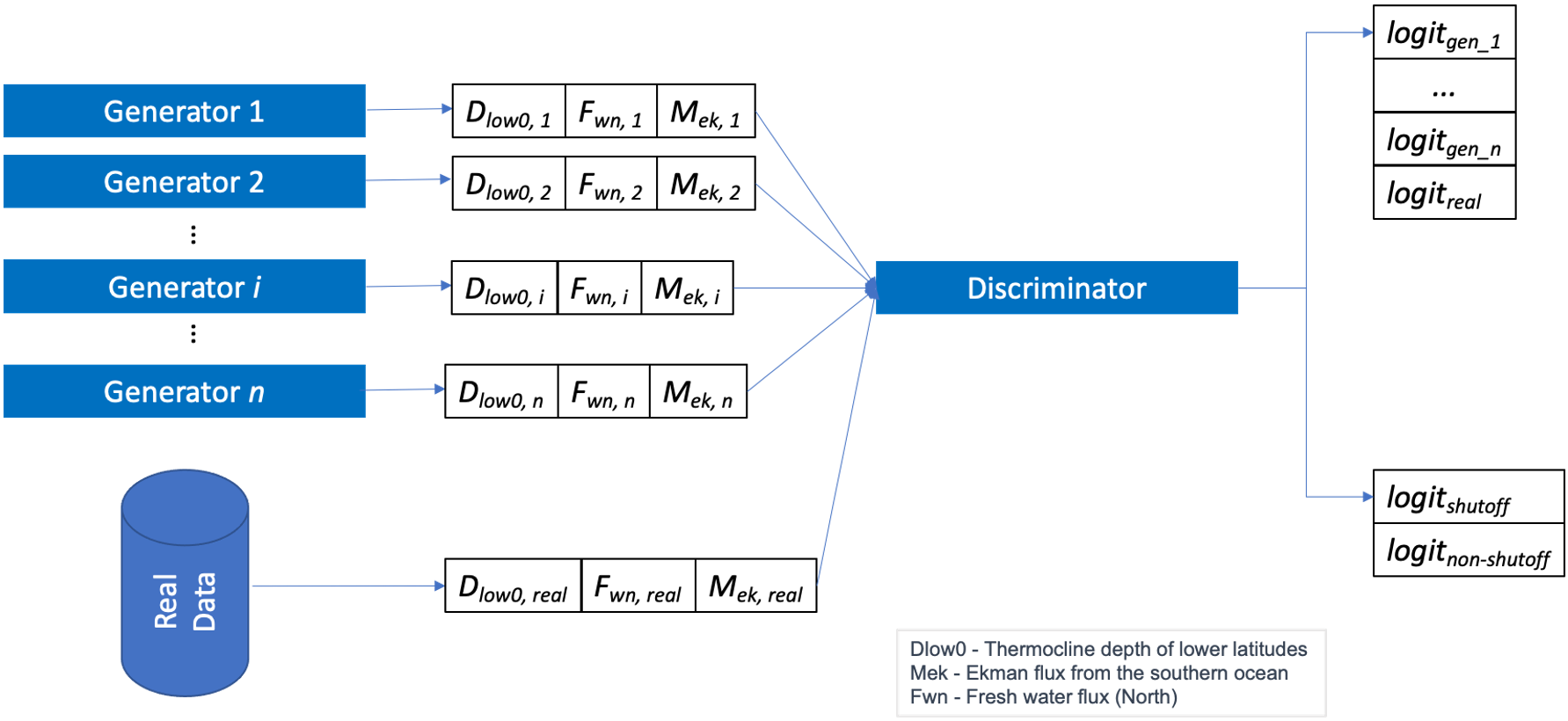}
\caption{A GAN Architecture that for Exploring the Area of Uncertainty.}
\label{fig:sep}
\end{figure}

To study the behavior of TIP-GAN with respect to learning the area of uncertainty specific to AMOC collapse, and roughly aligning with the unstable area in state space, or the separatrix, we used the four box model as the surrogate model.  We then reproduced one of the Gnanadesikan \cite{gnanadesikan2018flux} simulation experiments where three parameters were perturbed to study the AMOC overturning behavior. We show the TIP-GAN architecture for this experiment in Figure \ref{fig:sep}.

The goal of TIP-GAN in this experiment was to learn the boundaries of this area of AMOC instability (e.g. bifurcation region). We varied the number of generators from $N=(1,2,3)$ and built a GAN for each, resulting a total of three GANs.

\subsection{Recreating the Box Model Experiments Using TIP-GAN}

\begin{figure} 
\centering
\includegraphics[width=1\columnwidth]{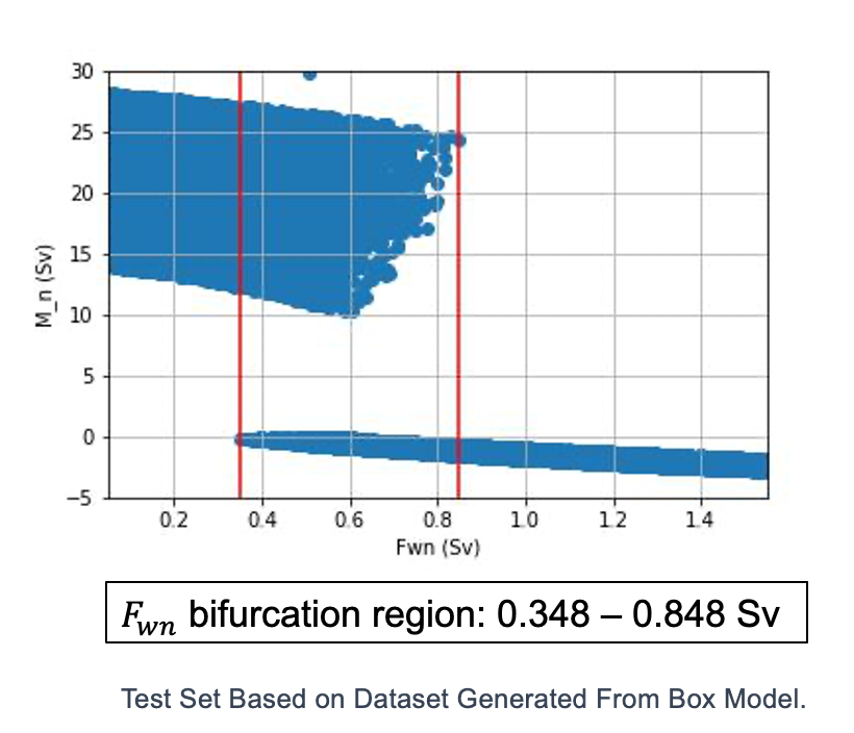}
\caption{The Final solutions for $M_n$ (vertical axis) vs. $F_w^n$ (horizontal axis) for 10,774 sets of parameters from Table \ref{tab:params}. Red lines show the Area of Uncertainty where multiple solutions are possible.}
\label{fig:uncert_test}
\end{figure}

The true dataset used for training and testing is built from uniformly sampling vectors of perturbed variables from a bounded 3-D subspace based on the four box model. The training dataset was composed of approximately 10,774 samples.  The test data set consisting 2,694 samples exploring how the uncertainty in Southern Ocean upwelling ($M_{ek}$) affected AMOC collapse. We generated initial ”on" and ”off" states by varying the initial depth of the low-latitude pycnocline. We then varied $M_{ek}$ between 15 and 35 Sv (comparable to the current range in climate models). Based on the Gnanadesikan experiments \cite{gnanadesikan2018flux}, we expect this to generate a family of curves with structure similar to those in Figures \ref{fig:fold} and \ref{fig:gan_sep}. We then allowed the freshwater flux to vary between 0.05 and 1.55 Sv, a wide enough range that for each value of $M_{ek}$ we could generate three domains, one with low $F_w^n$ where the overturning is always on (positive values, left-hand side of Figure \ref{fig:uncert_test}) a domain with high $F_w^n$ where it is always off (negative values, on right hand side of Figure \ref{fig:uncert_test} and an intermediate region of uncertainty (bounded by the red lines) in which whether we end up in an on or off state depends on our initial conditions. The perturbed parameters and their bounds are shown in  Table \ref{tab:params}.  All other variables were held constant.

\begin{table}
\resizebox{\columnwidth}{!}{
\begin{tabular}{||c c c||}
 \hline
 Parameter Name & Parameter Description & Bounds \\ [0.5ex] 
 \hline\hline
 $D_{low0}$ & Initial low latitude pycnocline depth (m) & [100.0, 400.0] \\ 
 \hline
 $M_{ek}$ & Ekman flux from the southern ocean (Sv) & [15, 35] \\
 \hline
 $F_{w}^n$ & Fresh water flux in North (Sv)  & [0.05, 1.55] \\
  \hline
\end{tabular}
}
\caption{Parameters that were perturbed for the Uncertainty Experiment.}

\label{tab:params}
\end{table}

We generated approximately 2,694 samples for each type of GAN (based on the number of generators) to measure its performance.  In Figure \ref{fig:uncert_test} we show the area of uncertainty between .348 Sv and .848 Sv, bounded by the two red lines.  As can be seen, the ”on" state samples and the ”off" state samples are consistent with the fold bifurcation states.  Again this carefully crafted experiment is calibrated to one of the experiments defined in the \cite{gnanadesikan2018flux} paper.

We evaluated the performance using the following evaluation metrics: 1. Percentage of generated samples within the bifurcation region and 2. Discriminator shutoff classification metrics (Precision, Recall, F1, Confusion Matrices).  We evaluated the generated sample for samples inside and outside of the bifurcation region.  This was compared to the test set and its number of samples inside and outside of this same region.

\section{Results and Analysis}
In this set of experiments we tried to answer a number of questions.  We wanted to better understand how increasing the number of generators affects the learning behavior.  We also wanted to better understand how incorporating discriminator uncertainty into the loss function would affect the outcome of learning.  More importantly we wanted to know if the GAN could discover input configurations that become more focused on the area of uncertainty over time.  In addition, we wanted to better understand how well the discriminator would perform in predicting ”on" and ”off" states after trained.

\begin{figure} 
\centering
\includegraphics[width=1\columnwidth]{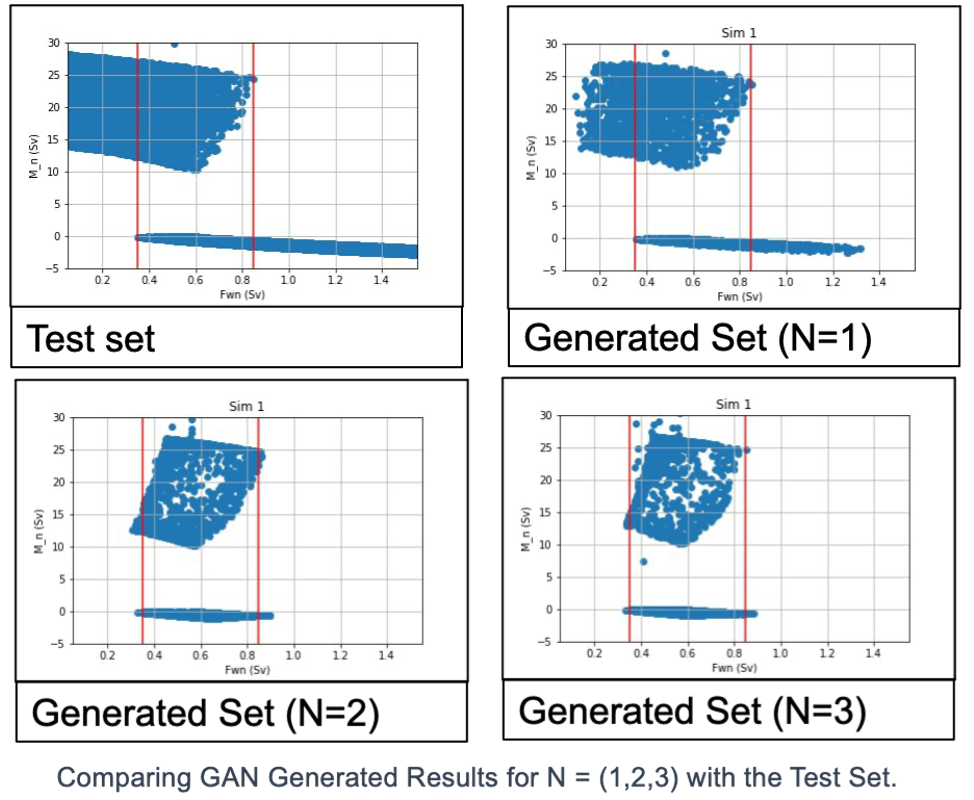}
\caption{Uncertainty and Increasing the Number of GANs.}
\label{fig:uncert_all}
\end{figure}

In Figure \ref{fig:uncert_all}, we show the results of this experiment.  We compare the samples generated from each GAN type (based on the number of generators used) and this area of uncertainty.  It appears that as we increase the number of generators the sampling is more confined this area of uncertainty.  The TIP-GAN appears to have learned the upper and lower bounds representing the final states lying along the bounding stable manifolds. The right-hand boundary of the upper point cloud represents cases that are close to the separatrix. The left-hand boundary of this point cloud represents end-states that began on the set of off-state stable manifolds.

In Table \ref{tab:uncertainty_regions} we show the percentage of samples for each dataset type and what percentage resides in this region of uncertainty, including the percentage of uncertainty for the training dataset, the percentage of uncertainty in the test dataset, and the percentage of uncertainty for each GAN with a set number of generators, where $N$ represents the number of generators.  The results show that the percentage of GAN-generated samples occurring within the region of uncertainty increases with increasing settings of $N$.

\begin{table}
\resizebox{\columnwidth}{!}{
\begin{tabular}{||c c c||}
 \hline
 Dataset Type & Number of Samples & Percent in Uncertainty Region  \\ [0.5ex] 
 \hline\hline
 Training & 10,774 & 34.9\% \\ 
 \hline
 Test & 2,694 & 35.5\% \\
 \hline
 TIP-GAN(N=1) & 2,694 & 67.4\%\\
  \hline
 TIP-GAN(N=2) & 2,694 & 91.4\%\\
  \hline
 TIP-GAN(N=3) & 2,694 & 98.7\%\\
  \hline
\end{tabular}
}
\caption{Comparing the Measured Amount of Uncertainty Learned by the Generators with the Training and Test Amounts.}
\label{tab:uncertainty_regions}
\end{table}

\begin{figure}[H]
\centering
\includegraphics[width=1\columnwidth]{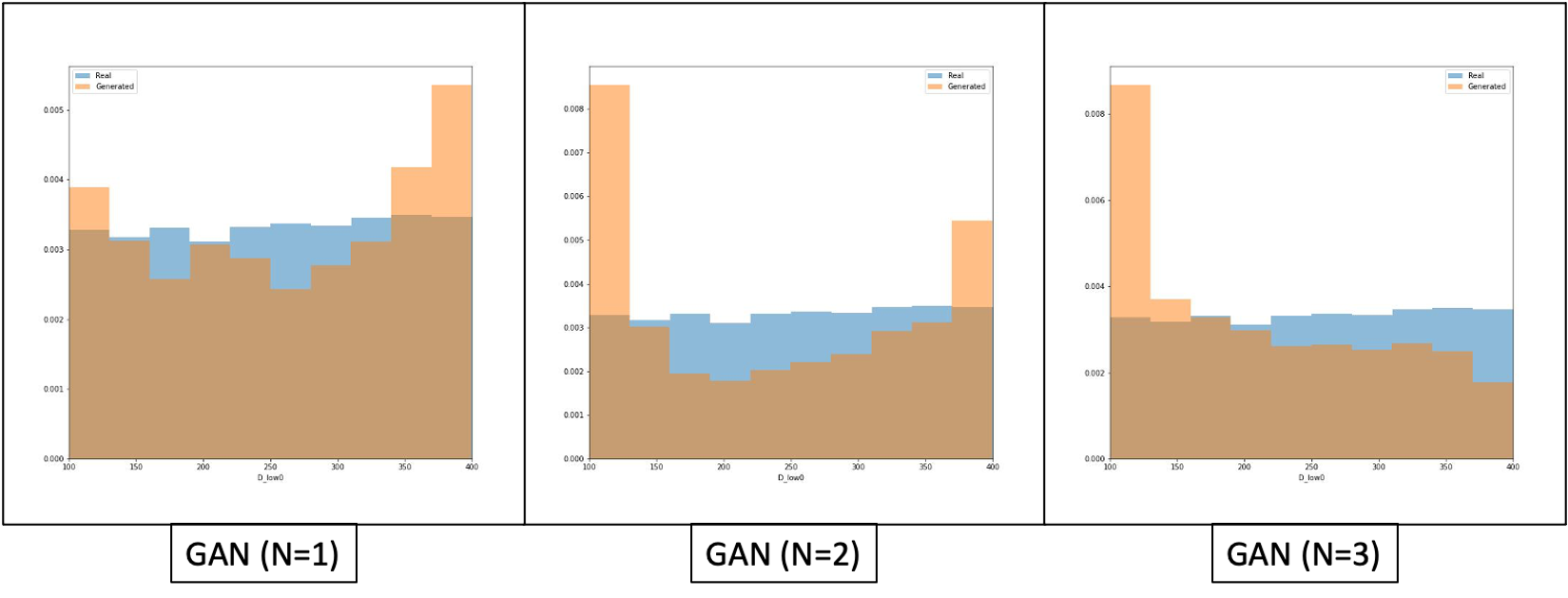}
\caption{Generator sampling distributions - $D_{low0}$}
\label{fig:dlow}
\end{figure}

\begin{figure}[H]
\centering
\includegraphics[width=1\columnwidth]{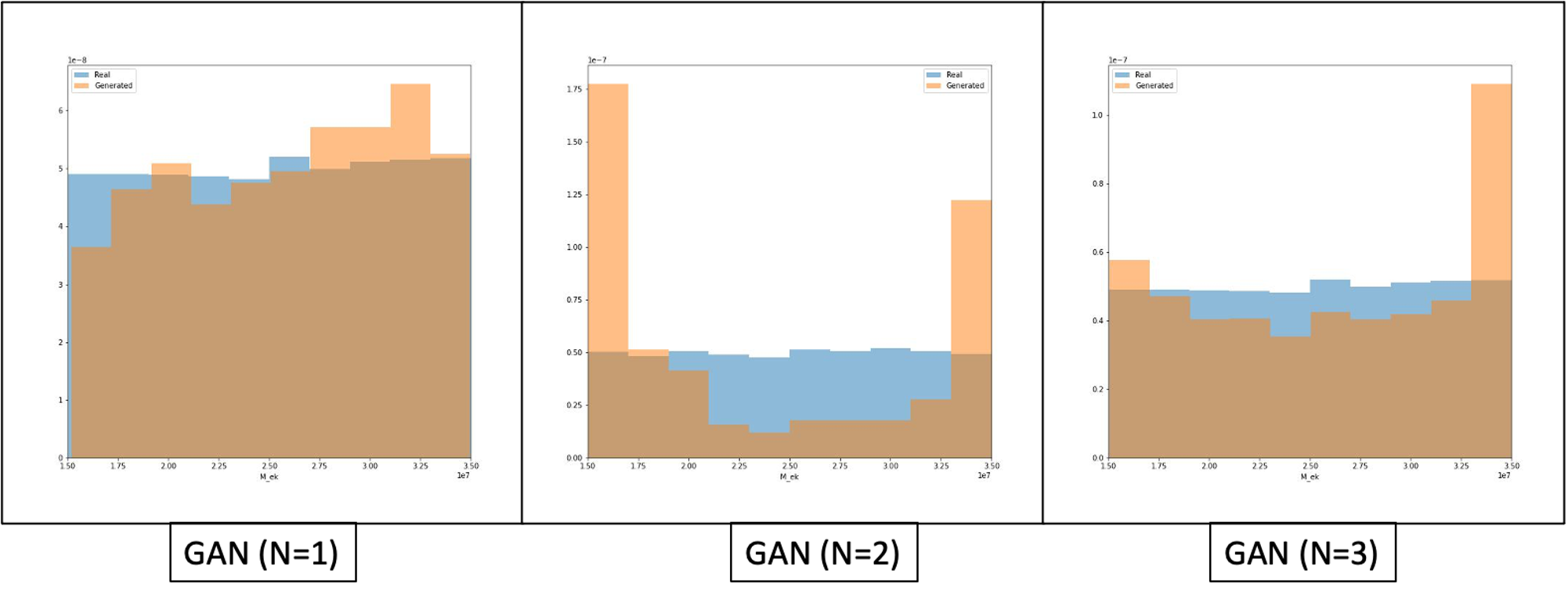}
\caption{Generator sampling distributions - $M_{ek}$
}
\label{fig:mek}
\end{figure}

\begin{figure}[H]
\centering
\includegraphics[width=1\columnwidth]{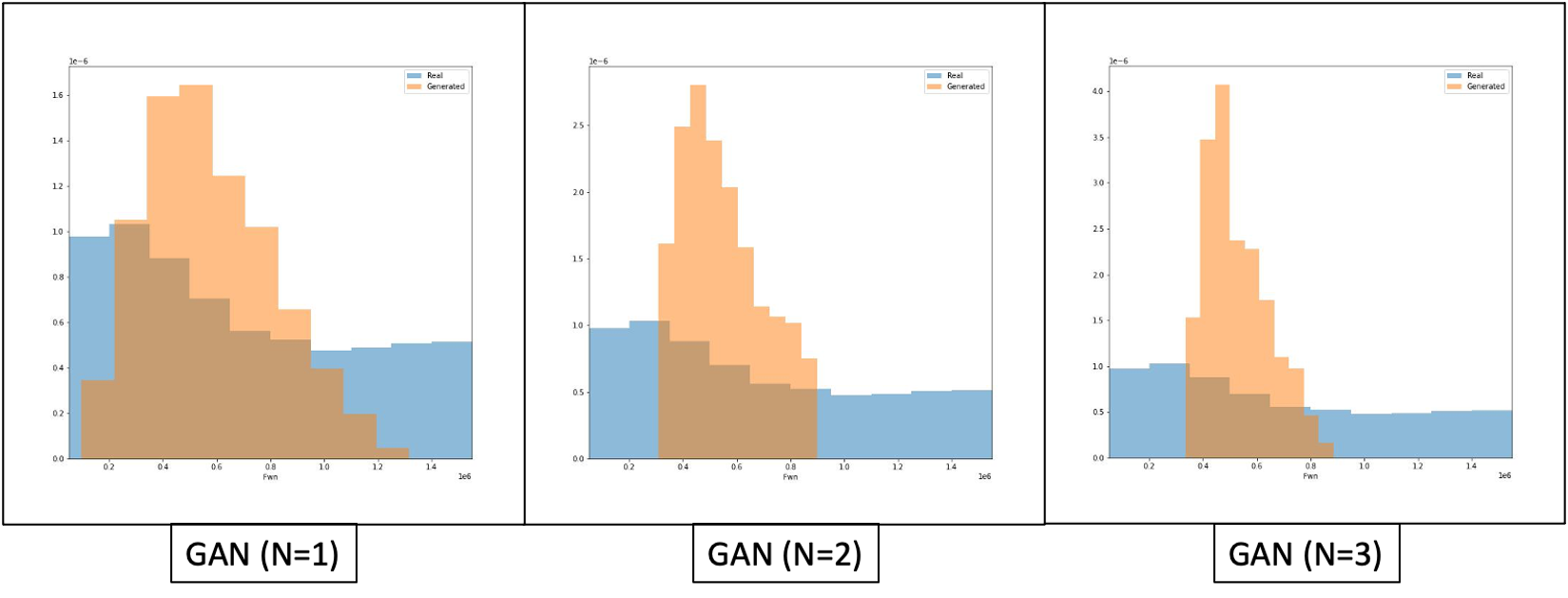}
\caption{Generator sampling distributions - $F_{w}^n$}
\label{fig:fwn}
\end{figure}

In Figures \ref{fig:dlow}, \ref{fig:mek}, and \ref{fig:fwn}, we show the distribution of sample generated as we increased the number of generators ($N$) for the three parameters perturbed  ($D_{low0}$, $M_{ek}$, and $F_{w}^n$), where the generated distribution is an overlay of the real distribution.


We show in Table \ref{tab:test_classifier} the precision, recall and F1 performance in terms of the test set.  We show these results in terms of the region of uncertainty vs. regions outside of the uncertainty.  The discriminator appears to perform marginally better on the normal region configurations vs. uncertainty region configurations.

\begin{table}[H]
\resizebox{\columnwidth}{!}{
\begin{tabular}{||c c c c c||}
 \hline
 & In Region & Precision & Recall & F1  \\ [0.5ex] 
 \hline\hline
 TIP-GAN (N=1) & Yes & 0.988 & 0.976 & 0.982 \\
 & No & 0.999 & 1.0 & 0.999 \\
 \hline
 TIP-GAN (N=2) & Yes & 0.995 & 0.995 & 0.995 \\
 & No & 1.0 & 1.0 & 1.0 \\
  \hline
 TIP-GAN(N=3) & Yes & 1.0 & 0.995 & 0.998 \\
 & No & 1.0 & 1.0 & 1.0 \\
  \hline
\end{tabular}
}
\caption{Test Classification Results.}
\label{tab:test_classifier}
\end{table}

\section{Future Work and Conclusions}
In this study we set out to demonstrate that a GAN could learn the area of uncertainty consistent with the separatrix in state space consistent with the four box model AMOC experiments.  Surprisingly when we allow the discriminator's uncertainty to influence the generator objective, as we increase the number of generators we see a more focused sampling on the area of uncertainty.  This result indicates that the TIP-GAN could likely be used to discover other types of bifurcations in state space and could be used for the original objective, as a way to guide the domain scientist (in this case the oceanographer) to areas in the parameter space that warrant a focused study for AMOC collapse.

Further work in this area is underway along three main lines. The first involves fitting the box model to a full climate model which has been found to exhibit rapid changes in the overturning circulation, following the work of \citet{levermann2010atlantic}. We will examine the extent to which the fitted model can explain/predict the thresholds at which these rapid changes occur. We will then run additional simulations in which we use the GAN to suggest parameter changes that push the model closer to a tipping point. The second line of research involves extending the box model to allow for overturning in the Pacific and Indian Oceans and using the GAN to examine the separatrix of this model in the presence of natural variability. Finally, we are working to connect the GAN with a neurosymbolic language to see whether we are able to go directly from natural language questions about the overturning to an optimal exploration of the state space.

\section{Acknowledgments}
Approved for public release; distribution is unlimited. This material is based upon work supported by the Defense Advanced Research Projects Agency (DARPA) under Agreement No. HR00112290032.

\bibliographystyle{aaai23.bst}
\bibliography{main}

\end{document}